\pdfoutput=1

\documentclass[11pt]{article}

\usepackage[preprint]{acl}

\usepackage{times}
\usepackage{latexsym}

\usepackage[T1]{fontenc}

\usepackage[utf8]{inputenc}

\usepackage{microtype}

\usepackage{inconsolata}

\usepackage[utf8]{inputenc} 
\usepackage[T1]{fontenc}    
\usepackage{hyperref}       
\usepackage{url}            
\usepackage{booktabs}       
\usepackage{amsfonts}       
\usepackage{nicefrac}       
\usepackage{microtype}      
\usepackage{xcolor}         

\usepackage{graphicx} 
\usepackage{wrapfig}
\usepackage[font=small,labelfont=bf]{caption} 
\usepackage{amsmath}
\usepackage{multirow}
\usepackage{algorithm}
\usepackage{multirow}
\usepackage{tabularx}
\usepackage{amsfonts}
\usepackage[most]{tcolorbox}

\tcbset{
  colback=gray!5!white,
  colframe=black!40!white,
  boxrule=0.4pt,
  arc=1mm,
  left=1mm,
  right=1mm,
  top=0.5mm,
  bottom=0.5mm,
  fonttitle=\bfseries,
  enhanced,
  sharp corners=south
}
\usepackage{algpseudocode}
\usepackage{float}
\usepackage{xcolor}  
\usepackage{arydshln}
\usepackage{cleveref}

\algrenewcommand{\algorithmiccomment}[1]{\hfill$\triangleright$ \textcolor{blue}{#1}}
\algrenewcommand{\alglinenumber}[1]{\footnotesize #1}
\algrenewcommand{\alglinenumber}[1]{\scriptsize #1}
\makeatletter
\renewcommand{\ALG@beginalgorithmic}{\small} 
\makeatother

\usepackage[most]{tcolorbox}
\usepackage{xcolor}

\definecolor{darkgreen}{RGB}{0,102,0}
\definecolor{darkred}{RGB}{180,0,0}

\definecolor{lightgraybox}{gray}{0.96}
\newcounter{takeaway}
\newtcolorbox{takeawaybox}[1][]{
    enhanced,
    colback=lightgraybox,
    colframe=black,
    boxrule=1pt,
    left=5pt, right=5pt, top=3pt, bottom=3pt,
    fonttitle=\bfseries,
    before skip=10pt, after skip=10pt,
    #1
}

\newcommand{\takeaway}[1]{%
    \refstepcounter{takeaway}%
    \begin{takeawaybox}
        \textbf{Takeaway \thetakeaway:}~#1
    \end{takeawaybox}
}

\usepackage{subfigure} 
\usepackage{enumitem} 
\newenvironment{itemize*}%
{\leftmargini=20pt\begin{itemize}%
\setlength{\itemsep}{3pt}%
\setlength{\parskip}{0pt}%
}%
{\end{itemize}} 
\newenvironment{enumerate*}%
{\begin{enumerate}%
\setlength{\itemsep}{0pt}%
\setlength{\parskip}{0pt}}%
{\end{enumerate}}

\NewDocumentCommand{\heng}
{ mO{} }{\textcolor{red}{\textsuperscript{\textit{Heng}}\textsf{\textbf{\small[#1]}}}}

\title{MAC: A Multi-Agent Framework for Interactive User Clarification in Multi-turn Conversations}

\author{
Emre Can Acikgoz$^{1}$, Jinoh Oh$^{2}$, Joo Hyuk Jeon$^{2}$, Jie Hao$^{2}$, \\
\textbf{Heng Ji$^{2}$, Dilek Hakkani-Tür$^{2}$, Gokhan Tur$^{2}$, Xiang Li$^{2}$, Chengyuan Ma$^{2}$, Xing Fan$^{2}$}\\
$^{1}$University of Illinois Urbana-Champaign, $^{2}$Amazon Alexa\\
\normalsize{\texttt{acikgoz2@illinois.edu},\; \texttt{ojino@amazon.com}}\\
}

\begin{document}
\maketitle
\begin{abstract}
  Conversational agents often encounter ambiguous user requests, requiring an effective clarification to successfully complete tasks. 
While recent advancements in real-world applications favor multi-agent architectures to manage complex conversational scenarios efficiently, ambiguity resolution remains a critical and underexplored challenge—particularly due to the difficulty of determining which agent should initiate a clarification and how agents should coordinate their actions when faced with uncertain or incomplete user input.
The fundamental questions of when to interrupt a user and how to formulate the optimal clarification query within the most optimal multi-agent settings remain open.
In this paper, we propose \textbf{MAC} (\textbf{M}ulti-\textbf{A}gent \textbf{C}larification), an interactive multi-agent framework specifically optimized to resolve user ambiguities by strategically managing clarification dialogues.
We first introduce a novel taxonomy categorizing user ambiguities to systematically guide clarification strategies. 
Then, we present MAC that autonomously coordinates multiple agents to interact synergistically with users. 
Empirical evaluations on MultiWOZ 2.4 demonstrate that enabling clarification at both levels increases task success rate 7.8\%  (54.5 $\rightarrow$ 62.3) and reduces the average number of dialogue turns (6.53 $\rightarrow$ 4.86) by eliciting all required user information up front and minimizing repetition.
Our findings highlight the importance of active user interaction and role-aware clarification for more reliable human–agent communication.

\end{abstract}

\section{Introduction}

Effective user clarification is fundamental to conversational agents, significantly impacting their ability to fulfill user requests accurately and efficiently~\citep{aliannejadi2019asking}. 
In natural interactions, users often express ambiguous queries, intentionally or unintentionally omitting details that seem inferable or contextually obvious. 
Such ambiguity can cause agents to make incorrect assumptions, provide incomplete responses, or even fail to accomplish tasks—issues that are especially critical in high-stakes domains such as healthcare, finance, and customer support.
Proactively resolving ambiguities through targeted user interactions by asking clear and relevant clarification questions can substantially enhance the accuracy of task execution, user satisfaction, and the overall effectiveness of conversational systems~\citep{deng2023proactivedialoguesurvey, acikgoz2025desideratum}.

\begin{figure}
  \centering
  \includegraphics[width=\linewidth]{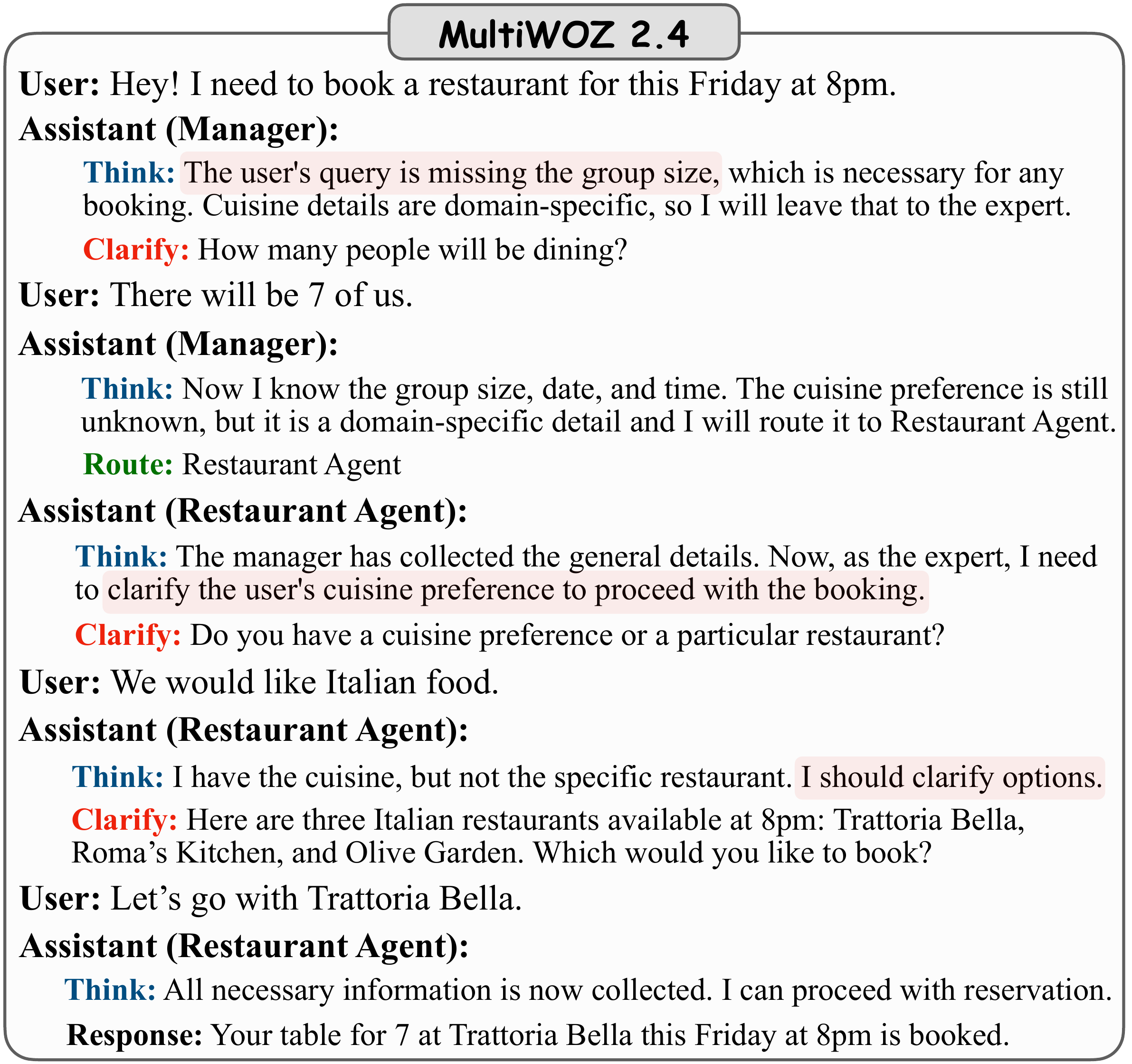} 
  \caption{\textbf{Multi-Agent Dialogue Illustrating Harmonized Clarification.} The manager agent first identifies and resolves only high-level, domain-agnostic ambiguity (group size), explicitly leaving domain-specific clarifications (such as cuisine and restaurant selection) to the domain expert which can require  domain knowledge from the database. This dialogue illustrates how each agent’s role is confined to its designated scope: the manager collects general requirements, while the expert gathers and confirms specialized details before completing the reservation.}
  \label{fig:fig1}
  \vspace{-10pt} 
\end{figure}

In single-agent systems, the challenge of ambiguity resolution has been previously studied with different strategies~\citep{dongre2024respact}, from asking targeted questions~\citep{li2023asktoclarify, zhang2025clarifywhennecessary} to inferring user preferences from past interactions~\citep{andukuri2024stargate}. However, the landscape of conversational AI is rapidly evolving towards more complex, multi-agent architectures, especially in industrial settings where a single agent cannot efficiently manage the large number of APIs and multitasking demands~\citep{sun2025multiagentappsurvey}. As a result, manager–expert routing systems are becoming the standard for handling real-world tasks~\citep{guo2024multiagentsurvey, tran2025multiagentmechnaism}. This paradigm, often featuring a "manager" or "advisory" agent that routes requests to specialized "expert" agents, introduces new layers of complexity for user interaction~\citep{ong2025routellm}. In such a setup, determining the optimal moment and method for clarification becomes a significant challenge. For instance, should the high-level advisory agent, which first receives the user's request, interrupt for clarification, or should this be delegated to a domain-specific expert agent, potentially increasing latency and conversational turns? Moreover, deciding how much domain-specific knowledge the manager should possess introduces another design challenge, establishing an essential knowledge boundary. Therefore, proposing an approach that effectively manages ambiguity resolution while remaining independent of this specific design choice is crucial for creating flexible and robust multi-agent systems. 

\begin{figure*}[t!]
    \centering
    \includegraphics[width=\textwidth]{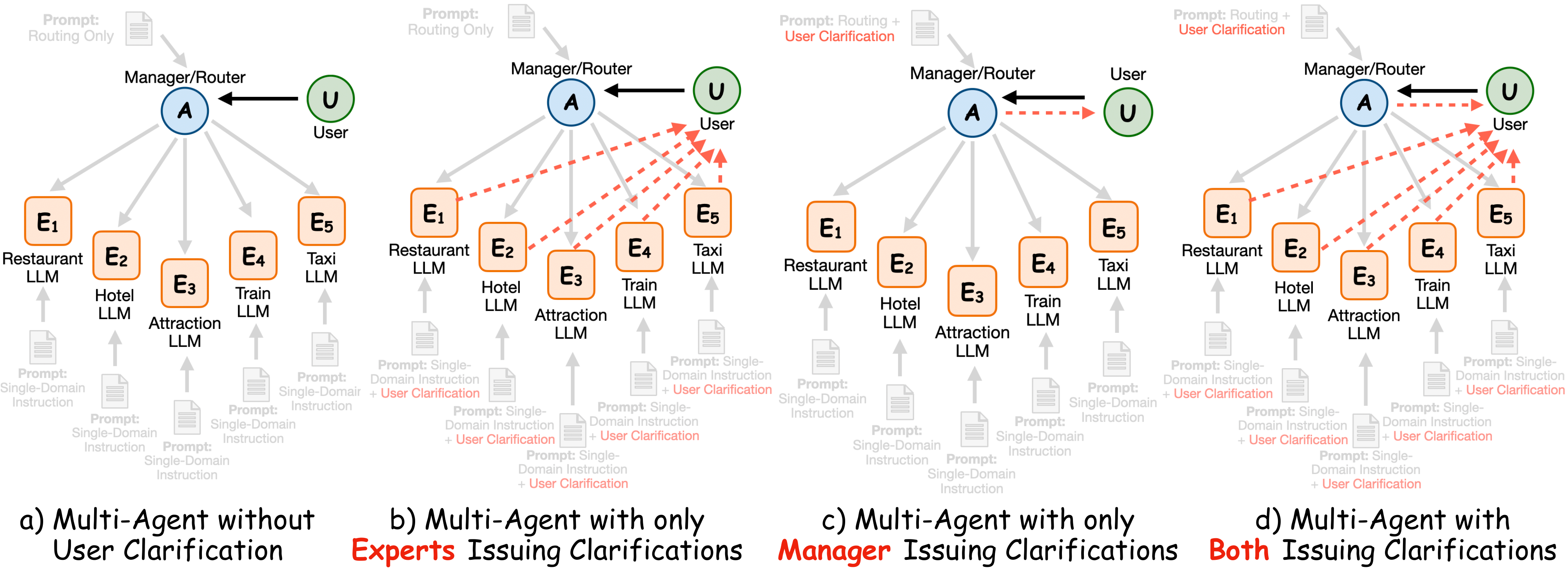}
    \vspace{-5mm}
    \caption{\textbf{A comparison of multi-agent architectures for user clarification in task-oriented dialogue.} \textbf{ (a) Baseline: }The manager/router agent routes user queries to domain-specific expert LLMs without any clarification. \textbf{(b) Experts-only clarification:} Each expert LLM independently interacts with the user to request clarification when needed, enhancing single-domain understanding. \textbf{(c) Manager-only clarification:} The manager/router agent requests clarification from the user before routing to any expert, enabling global disambiguation. \textbf{(d) Both:} Both the manager/router and the expert LLMs can independently interact with the user for clarification, allowing multi-level user-agent interaction. Dashed red arrows denote user clarification turns.}
    \vspace{-5mm}
    \label{fig:enter-label}
\end{figure*}

To explore these open questions, we introduce MAC (Multi-Agent Clarification), the first framework that focuses on resolving user ambiguities within multi-agent conversational systems and aims to uncover how, when, and by whom user clarifications should be initiated within these multi-agent settings (See \Cref{fig:fig1}). The framework strategically determines not only the moment to seek clarification but also which agent—the supervisor or the domain-specific expert—is best positioned to ask. Our experiments on MultiWOZ 2.4~\citep{ye2022multiwoz24} reveal that the placement and timing of clarification matter more than previously recognized. First, we show that enabling clarification at both levels delivers not just higher task success with a 7.8\% absolute gain over the no-clarification baseline, but does so while reducing the average number of conversational turns. Second, the optimal coordinated setup—where the supervisor manages high-level ambiguities and the expert agent resolves domain-specific ones—delivers the highest performance, even outperforming previous state-of-the-art TOD approaches on MultiWOZ with 11.50\%. This means effective clarification is not merely about “asking more questions”, but about delegating the right questions to the right agents at the right time.

The main contributions of this work are: 
\begin{itemize}[topsep=2pt, partopsep=-5pt, leftmargin=8pt, itemsep=-4.5pt] 
    \item We formalize the ambiguity resolution problem in \textbf{multi-agent conversational systems} with a taxonomy (\Cref{tab:ambig-taxonomy}), where decisions about \textit{when}, \textit{who}, and \textit{how} to clarify are jointly optimized among different agents.
    \item We propose \textbf{MAC}, the first multi-agent framework for user clarification, enabling distributed agents to dynamically coordinate clarification behavior.
    \item We show that coordinated clarification, when both manager and experts are empowered to ask targeted questions, leads to a 7.8\% absolute increase in task success (54.5\% $\rightarrow$ 62.3\%) while also reducing average dialogue length (6.53 $\rightarrow$ 4.86 turns) on MultiWOZ 2.4.
    \item We conduct extensive ablation studies, benchmarking MAC against strong single- and multi-agent baselines, analyzing the trade-offs of different clarification strategies, and demonstrating that our modular approach consistently outperforms prior TOD systems by a substantial margin. Additionally, we show that these gains are robust across diverse LLM backbones, including both proprietary and open-source models.
\end{itemize}

\section{Related Work}

\paragraph{Asking Clarification Questions}
Asking user clarification questions has been studied in conversational AI research, with distinct focuses on when to ask and what to ask~\citep{kuhn2022clam}. Some approaches use uncertainty estimation or information-theoretic models to decide when to initiate clarification~\citep{zhang2023clarify}. More advanced frameworks attempt to address both when and what to ask problems jointly~\citep{andukuri2024stargate, zhang2024modeling}, but they are often limited to a small number of conversational turns, which is insufficient for complex, real-world tasks. Closest to our approach, ReSpAct~\citep{dongre2024respact} enables clarification with rule-based prompting, however operating in vanilla single-agent settings. This approach fails to address the challenges of production systems, such as smart home platforms, which operate as complex multi-agent systems~\citep{guo2024multiagentsurvey}. Our work fills this critical gap by proposing a novel framework for coordinating clarification strategies across multiple specialized agents to ensure a coherent and efficient user experience similar to direct real world settings.

\paragraph{Large Language Models for Task-Oriented Dialogue}
Recent progress in LLMs has led to their adoption in multi-domain TOD systems~\citep{hudecek2023arellms}. Existing approaches typically rely on either prompting-based methods~\citep{hu2022icl, chung2023instructtod, xu2024autotod} or specialized fine-tuning~\citep{ehsan2020simpletod, yang2021ubar, zhong2023galaxy, sun2023mars, bang2023toatod, li2024fnctod}. Fine-tuned models are often tailored to narrow tasks such as state tracking or offline benchmarks, and as a result, they struggle to generalize to complex, real-world, multi-turn conversations~\citep{acikgoz2025coalm}. 
More recently, AutoTOD~\citep{xu2024autotod} demonstrated the use of GPT-4 with domain-specific, hand-crafted prompts and pre-defined APIs, but this approach depends heavily on lengthy instructions and lacks adaptability. On the other hand, some recent studies have begun to explore multi-agent architectures for TOD~\citep{gupta2024dard}, but they have not addressed the crucial aspect of user clarification, which is an essential skill for handling ambiguous or incomplete user requests in practical settings. In contrast, our work introduces the first multi-agent system explicitly focusing on asking user clarification question in multi-turn TOD, establishing an optimal framework for more reliable and user-centric dialogue agents.

\section{Environment}
\label{sec:dataset}

\begin{table*}[t!]
\centering
\resizebox{\textwidth}{!}{
\begin{tabular}{l|l|l}
\toprule
\textbf{Agent} & \multicolumn{1}{c|}{\textbf{Clarification Category}} & \multicolumn{1}{c}{\textbf{Description / Example}} \\
\midrule
\multirow{7}{*}{\textbf{Supervisor}} 
    & 1. Domain Ambiguity & User query could match multiple domains (e.g., \textit{“Find me a good place.”}) \\
    & 2. Intent Ambiguity & Domain is clear, but user’s goal is not (\textit{“Tell me about trains.”}) \\
    & 3. Vague Goal Specification & Query is too broad to act on (\textit{“Help me with my trip.”}) \\
    & 4. Contextual Disambiguation & Vague referents like “it” or “that place” are unclear \\
    & 5. General Conflict & Broad contradiction in user input (\textit{“I changed my mind about the date.”}) \\
    & 6. General Noise/Correction & Common errors/typos needing clarification (\textit{“I meant tomorrow not today.”}) \\
    & 7. Unfamiliar Domain Request & Request does not match any known domain (\textit{“Can you fix my phone?”}) \\
\midrule
\multirow{5}{*}{\textbf{Expert}}
    & 1. Parameter Underspecification & Missing key slot values (\textit{e.g., location, cuisine, people, time}) \\
    & 2. Value Ambiguity/Vagueness & Subjective terms require clarification (\textit{“a nice place”}) \\
    & 3. Constraint Conflict & Contradictory constraints (\textit{“a cheap but expensive restaurant”}) \\
    & 4. Entity Disambiguation/Not Found & Ambiguous or unrecognized entity (\textit{restaurant name not found}) \\
    & 5. Confirmation of Inferred Information & Inferred detail from context needs user confirmation \\
\bottomrule
\end{tabular}
}
\caption{\textbf{Clarification Taxonomies for Ambiguity Handling: Supervisor and Experts.} Supervisor agent addresses only high-level, domain-agnostic ambiguities, while the Expert agent resolves domain-specific underspecification prior to API execution.}

\label{tab:ambig-taxonomy}
\end{table*}

\paragraph{MultiWOZ 2.4} In our research, we utilize MultiWOZ 2.4~\citep{ye2022multiwoz24}, a comprehensive multi-domain dialogue benchmark designed for task-oriented dialogue (TOD) systems. It contains multi-turn conversations between a user and a system simulating a dialogue assistant. The user is given a goal (e.g., book a hotel and a restaurant in the same area), while the system must fulfill the request using a consistent belief state and database API. Each dialogue is annotated with dialogue acts, belief states, and system actions at each turn, enabling full end-to-end modeling and evaluation. It consists of $\sim$8,500 training dialogues and a test set of 1,000 dialogues. The test conversations within MultiWOZ 2.4 simulate customer service interactions across five distinct domains: restaurant, hotel, train, attraction, and taxi. To create more realistic real-world scenarios in multi-turn settings with actual users, we enhanced this dataset by incorporating a user simulator, making the tasks both more challenging and more authentic.

\paragraph{Task} During evaluation, each dialogue involves between up to five domains: \textit{restaurant}, \textit{hotel}, \textit{train}, \textit{attraction}, and \textit{taxi}. The agent must understand the user’s multi-intent goals, track the evolving belief state, issue database queries, and generate appropriate system responses. Crucially, user queries often underspecify constraints (e.g., "Book a restaurant for dinner"), requiring the agent to proactively request clarification (e.g., number of people, cuisine, or time), which makes a well-suited environment to test our multi-agent approach.

\paragraph{User Simulator}  Our experimental setup involves a user simulator, which we implemented to interact with the agent in a multi-turn conversational flow. This simulator is tasked with pursuing the predefined user goals, while the agent's objective is to assist the user in achieving these goals by interacting with an external database. We chose to work with MultiWOZ 2.4 over other benchmarks, due to its capability to simultaneously handle five distinct domains within a single conversational context. The available actions for each domain are detailed in Appendix \Cref{tab:dataset}.






\begin{algorithm*}[h]
\caption{MAC: Multi-Agent Clarification Workflow}
\label{alg:mac}
\begin{algorithmic}[1]
\Require User query $u_t$ at dialogue turn $t$; supervisor agent $A_S$; domain experts $A_E=\{A_{d_1},\dots,A_{d_n}\}$
\Ensure Either \textsc{Clarify}$(q)$ or \textsc{Respond}$(r)$

\Function{MAC}{$u_t$}
    \If{$A_S.\text{is\_ambiguous}(u_t)$}
        \State $q \gets A_S.\text{ask\_clarification}(u_t)$
        \State \Return \textsc{Clarify}$(q)$ \Comment{Supervisor requests disambiguation}
    \EndIf

    \State $d \gets A_S.\text{select\_domain}(u_t)$
    \State $A_d \gets \text{the expert for domain } d$ \Comment{Route to best-fit expert}

    \If{$A_d.\text{is\_ambiguous}(u_t)$}
        \State $q_d \gets A_d.\text{ask\_clarification}(u_t)$
        \State \Return \textsc{Clarify}$(q_d)$ \Comment{Expert requests a targeted follow-up}
    \Else
        \State $r \gets A_d.\text{execute\_domain\_response}(u_t)$
        \State \Return \textsc{Respond}$(r)$ \Comment{Final, domain-grounded answer}
    \EndIf
\EndFunction

\vspace{0.25em}
\Statex \textbf{Design notes.} \textcolor{blue}{(i) Only one clarification is issued per turn to limit latency. (ii) $A_S$ handles global ambiguity; $A_d$ handles domain-specific gaps. (iii) Routing uses $A_S.\text{select\_domain}$, which may rely on intent classification or retrieval over domain schemas.}
\end{algorithmic}
\end{algorithm*}
\section{Method}

Ambiguity in user requests is a central challenge in conversational agents, yet there is limited empirical guidance in multi-agent dialogue systems on where and how clarification should be initiated in such frameworks. To address this, we systematically investigate different agent-level strategies for user clarification in a hierarchical multi-agent architecture comprising a manager/router and multiple domain-specific experts (See \Cref{fig:enter-label}).

\subsection{MAC: Multi-Agent Clarification for User Ambiguities}
In our multi-agent framework, we adopt a centralized multi-agent setting as our base architecture, which consists of a single \textbf{supervisor agent} and multiple specialized \textbf{domain expert agents}. The supervisor agent is responsible for orchestrating the overall dialogue flow by routing each user request to the most relevant domain expert, following the router-based approach in~\citet{ong2025routellm}. Each expert agent is specialized for one of the five domains in MultiWOZ 2.4, and is tasked with executing domain-specific actions to fulfill user goals. In MAC, we further enhance this framework by integrating user clarification mechanisms to resolve ambiguities. This involves assigning specific clarification-handling capabilities to both supervisor and domain expert roles as in \Cref{tab:ambig-taxonomy}, enabling them to manage different forms of uncertainty and improve final task outcomes.

\paragraph{Supervisor Agent}
In the MAC framework, the Supervisor agent is responsible for two different tasks: (i) orchestrating the agent collaboration by routing user queries to the appropriate domain expert, and (ii) handling top-level clarification of user requests when the ambiguity can be resolved with general commonsense reasoning, independent of domain-specific knowledge (\Cref{tab:ambig-taxonomy}, top). Formally, for each incoming user query $u$, the Supervisor evaluates an ambiguity function $\text{is\_ambiguous}(u) \in {0,1}$: if $\text{is\_ambiguous}(u) = 1$, the agent issues a clarification prompt to the user using the standardized format \texttt{<clarify>{question}</clarify>}; otherwise, it selects the appropriate domain expert with \texttt{<route>{domain}</route>}. This prompt-based control flow is illustrated in \Cref{fig:mac-advisor}, where the Supervisor's output is parsed and dispatched to downstream agents. Notably, the Supervisor operates without access to domain-specific databases or APIs, ensuring that only non-domain-specific ambiguities (e.g., group size or intent) are addressed at this stage. After resolving high-level ambiguities, supervisor delegates the (potentially clarified) user request to the corresponding domain expert, enabling more efficient and role-aware collaboration across the agent hierarchy. 

\paragraph{Domain Expert Agents}

Each Domain Expert agent is responsible for executing user goals within a specific task domain. We instantiate five expert agents, corresponding to the five domains in MultiWOZ 2.4: \texttt{restaurant}, \texttt{hotel}, \texttt{train}, \texttt{taxi}, and \texttt{attraction}. Once a user query is routed to a domain expert, the agent analyzes the input—potentially enriched by prior supervisor-level clarification—and determines whether the information is sufficient to proceed with an \textit{accurate API calls} or \textit{reliable response generation}. To guide this behavior, we prompt each expert individually with domain-specific instructions that are coupled with the standardized protocols for user clarification (see \Cref{fig:mac-expert}), following predefined expert specific clarification taxonomy (\Cref{tab:ambig-taxonomy}, bottom). Similar to the supervisor, the agent computes an ambiguity function $\text{is\_ambiguous}(u) \in {0,1}$; if the result is $1$, the agent triggers a clarification request formatted as \texttt{<clarify>{question}</clarify>}. If the input is deemed sufficient, the agent executes the necessary domain-specific operations and responds using the structure \texttt{<response>{utterance}</response>}.Otherwise, the agent is allowed to ask multiple clarification questions until the total conversation length exceeds 20 turns. These prompt-structured outputs allow the framework to dynamically interleave reasoning, clarification, and execution in multi-turn interactions. Domain Experts have access to API schemas and databases corresponding to their domain, enabling them to ground their responses in task-specific constraints and complete user requests accurately. 

To elucidate the core principles of MAC, we conduct a systematic analysis of different strategies across the experimental design choices detailed in \Cref{fig:enter-label}.

\section{Experiments and Results}

\begin{table*}[t]
\centering
\resizebox{\textwidth}{!}{%
\begin{tabular}{lccccc}
\toprule
\textbf{Method}        & \textbf{Clarification} & \textbf{Success (Max@5 $\uparrow$)} & \textbf{Success (Avg@5 $\uparrow$)} & \textbf{Avg. Turns ($\downarrow$)} \\ \midrule
MAC w/o Clarification  & -                      & 54.5                                & 53.72 ± 0.92                         & 6.53 \\
MAC$_{expert}$         & Expert                 & 55.6                                & 54.88 ± 1.04                         & 5.53 \\
MAC$_{supervisor}$     & Supervisor             & 57.1                                & 55.50 ± 1.86                       & 5.11 \\ \hdashline[0.5pt/2pt] \addlinespace[1mm]
MAC                    & Both                   & \textbf{62.3}                       & \textbf{58.40 ± 2.10}                & \textbf{4.86} \\ \bottomrule
\end{tabular}%
}
\caption{\textbf{Main results on MultiWOZ 2.4.} Main results comparing different prompting and clarification strategies in the MAC framework on MultiWOZ 2.4. We report (\textbf{Success Max@5}): the highest single-run task success rate out of five runs, (\textbf{Success Avg@5}): the mean and standard deviation of success rates over five runs, and (\textbf{Avg. Turns}): the average number of dialogue turns per conversation (lower is better). Each row corresponds to a specific agent configuration—clarification enabled for the expert, the supervisor, both, or neither. Results demonstrate that enabling clarification for both supervisor and expert agents leads to the highest task success and most efficient dialogues.
}

\label{tab:main-results}
\end{table*}

\subsection{Experimental Setup}
In our MAC framework, we used \texttt{gpt-4o-2024-08-06} as the base configuration for the selected LLM, serving as both the advisor and each expert, unless otherwise specified. However, we have conducted comprehensive ablation studies on the effect of model choice for both nodes in the \Cref{subsec:ablation}. We conducted our evaluation on the MultiWOZ 2.4 test split, which contains 1,000 test samples from five domains: restaurant, hotel, train, attraction, and taxi. The evaluation was performed in online sessions where we implemented a user simulator based on \texttt{gpt-4o-2024-08-06}, as defined in \Cref{sec:dataset}. To account for LLM randomness, we ran each experiment five times and report the Success Rate with Avg@5 with their standard deviations and also include Success Rate with Max@5 which gives the max scores achieved in these five runs. Further details about the evaluation metrics can be seen in \Cref{app:evaluation}.

\subsection{Baselines}
MAC is the first LLM-based multi-agent framework specifically designed for user clarification. To evaluate its effectiveness, we compare it against three variants of the same multi-agent architecture: (i) without any user clarification (see \Cref{fig:supervisor-prompt,fig:expert-prompt} for baseline supervisor and expert prompts without clarification), (ii) with user clarification handled only by the Supervisor, and (iii) with user clarification enabled only for the domain experts (see \Cref{fig:enter-label}). In setting (i), neither the Supervisor nor the domain experts are instructed to ask clarification questions. In setting (ii), only the Supervisor is prompted to perform both routing and user clarification, while the domain experts are limited to responding after routing. In setting (iii), only the domain experts are prompted to ask clarification questions, and the Supervisor is responsible solely for routing. In contrast, MAC enables user clarification at both the Supervisor and domain expert levels, allowing every agent node to interact with the user as needed. This setup allows for a fair and systematic evaluation of the individual and combined effects of clarification skills across different nodes.

\subsection{Main Results}
We compare MAC against three variants: (i) MAC without any clarification capability, (ii) MAC where only domain-specific experts perform clarifications (MAC$_{expert}$), and (iii) MAC where only the supervisor at the top node initiates clarification questions (MAC$_{supervisor}$). \Cref{tab:main-results} summarizes our main findings, demonstrating the effectiveness of the proposed MAC framework. Specifically, our proposed MAC framework achieves an increase in task accuracy of approximately 8\% at maximum and about 5\% on average compared to the no-clarification baseline. Remarkably, this improvement is accompanied by a reduction in the average dialogue length by roughly two conversational turns. Our results clearly indicate that prompting agents to proactively clarify ambiguous, incomplete, or underspecified user requests consistently improves task success rates without extending dialogue length. This emphasizes the benefit of proactive conversational strategies. These findings highlight the MAC framework’s superior performance, effectively balancing accuracy with conversational efficiency.

\takeaway{Asking clarification questions consistently \textbf{increase task success} and \textbf{decrease number of turns} to solve the task in multi-agent settings.}

\subsection{Ablation Studies}
\label{subsec:ablation}

\paragraph{Comparison of MAC with Other TOD Systems}


\begin{table}[t]
\centering
\resizebox{\columnwidth}{!}{%
\begin{tabular}{lc}
\toprule
\textbf{Method}                     & \textbf{Success Rate ($\uparrow$)} \\ \midrule
SimpleTOD~\cite{ehsan2020simpletod} & 22.00 \\
UBAR~\cite{yang2021ubar}            & 26.80 \\
GALAXY~\cite{zhong2023galaxy}       & 28.80 \\
MARS~\cite{sun2023mars}             & 27.50 \\
TOATOD~\cite{bang2023toatod}        & 26.90 \\
FNCTOD~\cite{li2024fnctod}          & 44.40 \\
AutoTOD~\cite{xu2024autotod}        & 46.90 \\ \hdashline[0.5pt/2pt] \addlinespace[1mm]
MAC                                 & \textbf{58.40} \\ \bottomrule
\end{tabular}%
}
\caption{\textbf{MAC's Performance Compared to Existing TOD Systems.} Evaluation of various TOD methods using a standardized framework. Results for baseline models are sourced from AutoTOD to ensure fair and consistent comparison. Results taken from \cite{xu2024autotod}, following same evaluation protocol to ensure fairness in our comparison.}
\vspace{-5mm}
\label{tab:mac-tod}
\end{table}
In \Cref{tab:main-results}, we demonstrated MAC’s performance compared to its variants, highlighting that combining clarification capabilities between supervisor and experts results in the most effective setup. To further contextualize MAC's performance, it is crucial to benchmark against other leading task-oriented dialogue (TOD) systems. Following the evaluation framework of AutoTOD~\citep{xu2024autotod}, we present this comparison in \Cref{tab:mac-tod}. Our results indicate that MAC surpasses previous state-of-the-art models, achieving an improvement of approximately 11.50\% over the closest agent AutoTOD. This underscores both the robustness of the multi-agent architecture in multi-domain scenarios such as MultiWOZ 2.4 and the critical role of proactive clarification when handling uncertainties~\footnote{Some earlier TOD systems in \Cref{tab:main-results} were developed prior to the integration of LLMs and follow fundamentally different pipelines~\citep{acikgoz2025todllms}, making direct comparisons not fully fair. Nevertheless, this comparison aims to contextualize MAC’s performance and also illustrate the overall progress of TOD systems over time.}. 

\takeaway{MAC demonstrates \textbf{superior performance over previous TOD systems}, attributed to its multi-agent architecture and effective user clarification capabilities.}



\begin{table}
\centering
\resizebox{\linewidth}{!}{%
\begin{tabular}{lc}
\toprule
\textbf{Language Model} & \textbf{Success Rate ($\uparrow$)} \\
\midrule
\multicolumn{2}{l}{\textbf{MAC w/o Clarification}} \\
\hdashline[0.5pt/2pt] \addlinespace[1mm]
\hspace{3mm}\texttt{gpt-4o} & 53.72 ± 0.92 \\
\hspace{3mm}\texttt{gpt-4o-mini} & 52.40 ± 2.08 \\
\hspace{3mm}\texttt{Qwen3-235B-A22B} & 47.32 ± 1.72 \\
\cmidrule(lr){1-2}
\multicolumn{2}{l}{\textbf{MAC with Clarification}} \\
\hdashline[0.5pt/2pt] \addlinespace[1mm]
\hspace{3mm}\texttt{gpt-4o} & 58.40 ± 2.10 (\textcolor{darkgreen}{+4.68}) \\
\hspace{3mm}\texttt{gpt-4o-mini} & 57.10 ± 1.42 (\textcolor{darkgreen}{+4.70}) \\
\hspace{3mm}\texttt{Qwen3-235B-A22B} & 54.50 ± 1.06 (\textcolor{darkgreen}{+7.28}) \\
\bottomrule
\end{tabular}%
}
\caption{MAC success rates with different LLMs, with and without clarification. \textcolor{darkgreen}{\textbf{Values in parentheses}} show absolute improvement from clarification.}
\vspace{-5mm}
\label{tab:mac-llm-results}
\end{table}

\paragraph{How does the choice LLM effect MAC?}
Since multi-agent setups are typically constructed using multiple LLMs with prompting, it is valuable to evaluate the performance of diverse LLMs within our MAC framework.
To this end, we experimented with proprietary API-based and open-source models: \texttt{GPT-4o-2024-11-20}~\citep{hurst2024gpt4o}, \texttt{gpt-4o-mini}, \texttt{Qwen3-235B-A22B}~\citep{yang2025qwen3}. As shown in \Cref{tab:mac-llm-results}, enabling coordinated user clarification for both supervisor and expert agents in the MAC framework consistently improves task success rates, regardless of model type. For instance, \texttt{gpt-4o} and \texttt{gpt-4o-mini} achieve absolute improvements of +4.68 and +4.70 points, respectively, when equipped with clarification. 
Notably, the open-source \texttt{Qwen3-235B-A22B} model exhibits an even larger gain of +7.18 points, narrowing the gap with proprietary counterparts. The larger delta in accuracy for open-source LLMs suggests that well-designed supervision and agent coordination can unlock their potential, making them competitive candidates with propriety models for agentic systems in practice.

\takeaway{Enabling user clarification for both supervisor and expert agents in MAC consistently improves performance \textbf{regardless of model type}, even with open-source models.}

\takeaway{Proactive interaction and effective agent coordination yield the highest accuracy gains for open-source LLMs, making them as strong alternatives to proprietary models in agentic systems.}
 \vspace{-3mm}

\begin{figure}[h!]
    \centering
    \includegraphics[width=\linewidth]{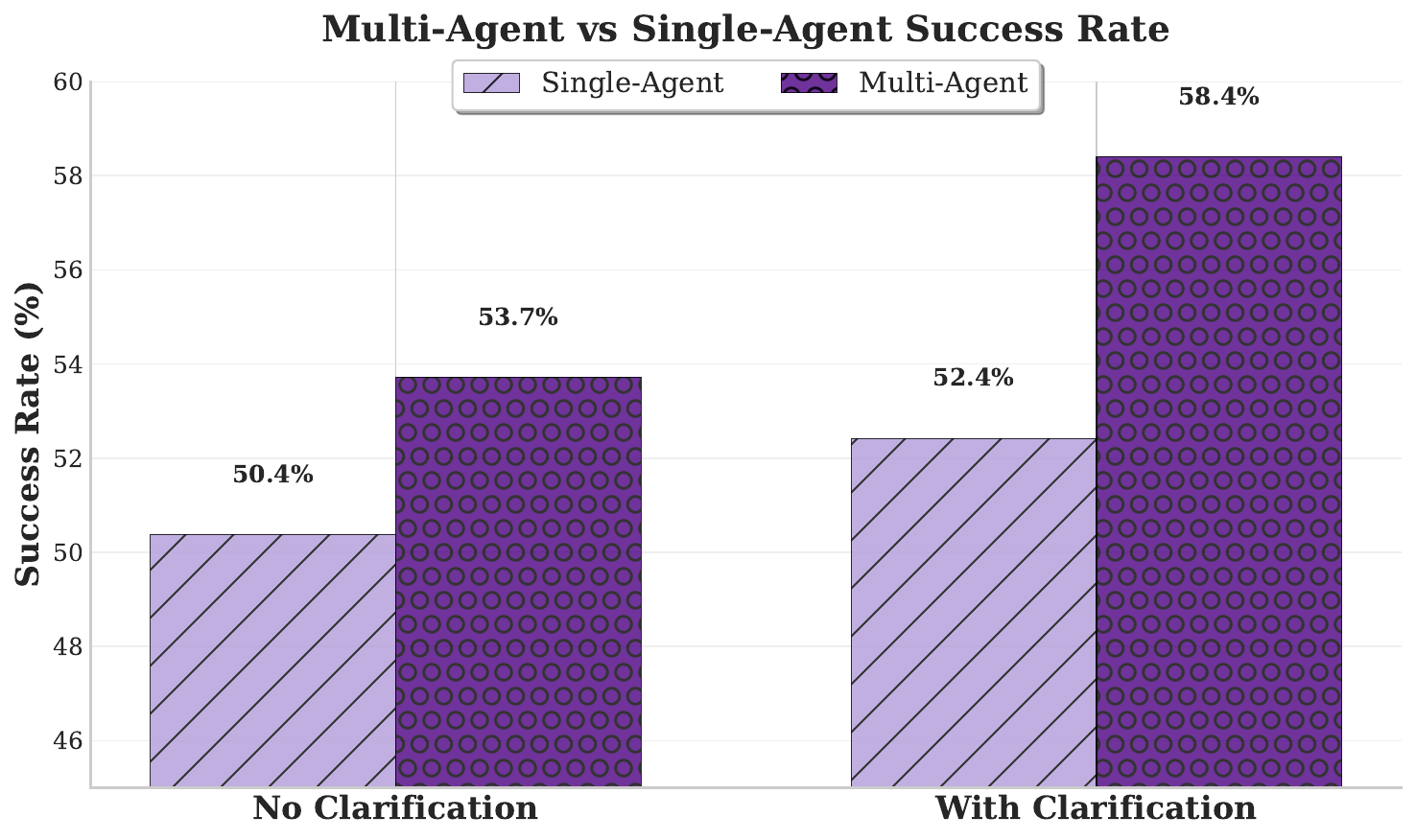}
    \caption{\textbf{Comparison of Single-Agent vs Multi-Agent Systems With and Without Clarification.} The multi-agent approach benefits more from user clarification, achieving the highest performance.}
    \vspace{-3mm}
    \label{fig:mac-sac}
\end{figure}

\subsection{MAC vs Single Agent Clarification}

\Cref{fig:mac-sac} presents a comparative analysis of single-agent and multi-agent systems using \texttt{GPT-4o-2024-11-20}, examining their performance with and without user clarification. Our findings demonstrate that clarification enhances success rates in both setups; however, the improvement is notably more pronounced in the multi-agent configuration. Specifically, the Multi-Agent Clarification (MAC) system outperforms Single-Agent Clarification (SAC) by 6\% (52.4\% $\rightarrow$ 58.4\%), highlighting the advantage of separating responsibilities between a Supervisor, responsible for general, high-level ambiguities, and domain-specific Experts who handle specialized clarifications. Moreover, our multi-agent setup consistently achieves higher success rates than the single-agent approach even in scenarios without clarification, emphasizing the inherent benefit of distributing workload among coordinated agents rather than overloading a single agent with multiple roles. This superior performance, particularly under clarification conditions, underscores the advantages of modularity in resolving ambiguities effectively and efficiently scaling to complex, multi-domain interactions. In contrast, single-agent systems face diminishing returns and decreased interpretability as complexity grows due to the necessity of internalizing diverse expertise. Additionally, the modular structure of MAC facilitates incremental updates and streamlined integration of new domains, enhancing its robustness and practical applicability in real-world scenarios, as demonstrated by systems such as the model context protocol (MCP)~\citep{hou2025mcp}. 

\takeaway{Multi-agent setup outperforms single-agent both with and without clarification, which suggest \textbf{multi-agent setup offers improved scalability and efficiency} as each agent operates with a more focused and concise context.}

\paragraph{Effect of Clarification Taxonomies on Task Success}
\begin{table}[h!]
\centering
\resizebox{\columnwidth}{!}{%
\begin{tabular}{lc}
\toprule
\textbf{Method} & \textbf{Success Rate ($\uparrow$)} \\
\midrule
MAC & 58.40 ± 2.10 \\
\hdashline[0.5pt/2pt] \addlinespace[1mm]
\hspace{3mm} w/o Ambiguity \& Vagueness Handling 
    & 52.20 ± 1.21 (\textcolor{darkred}{-6.20}) \\
\hspace{3mm} w/o Slot/Parameter-Blocking Clarification 
    & 56.22 ± 1.36 (\textcolor{darkred}{-2.18}) \\
\bottomrule
\end{tabular}%
}
\caption{Ablation study on the impact of supervisor (Ambiguity \& Vagueness Handling) and expert (Slot/Parameter-Blocking Clarification) clarification taxonomies, with the drop in accuracy (\textcolor{darkred}{\textbf{red}}).}
\label{tab:taxonomy-ablation}
\end{table}

To better understand the critical roles of different clarification strategies and understand the correspondence between the different ambiguity types defined in the taxonomy, we clustered the supervisor’s taxonomy into high-level Ambiguity and Vagueness Handling (encompassing Domain Ambiguity, Intent Ambiguity, Vague Goal Specification, and Contextual Disambiguation) and the expert’s taxonomy into Slot/Parameter Uncertainty (covering Parameter Underspecification and Value Ambiguity/Vagueness). As shown in \Cref{tab:taxonomy-ablation} based on \texttt{GPT-4o-2024-11-20} outputs, ablating Ambiguity and Vagueness Handling from the supervisor yields a substantial drop in task success rate ($-$6.20), highlighting the importance of proactively resolving common-sense ambiguities before delegating to domain-specific experts. In contrast, removing Slot/Parameter-Blocking Clarification from experts causes a smaller, but still notable, accuracy decline ($-$2.18), confirming that careful handling of underspecified or vague user slots is also essential, especially for API calls requiring precise parameters. It demonstrates that both forms of clarification are necessary, but high-level disambiguation with supervisor is particularly crucial for robust multi-agent dialogue.

\takeaway{High-level \textbf{ambiguity and vagueness handling by the supervisor is essential} for robust performance, with its removal causing the largest drop in task success among all clarification skills.}

Overall, these results support our preference for the multi-agent architecture, where MAC delivers higher accuracy, improved scalability, and enhanced maintainability, underscoring the critical role of modular design in developing robust and generalizable conversational agents.



\section{Discussion}

\paragraph{Conclusions}
We introduce MAC, the first multi-agent LLM framework specifically designed for interactive user clarification in conversational agents, and we also present the first comprehensive user clarification taxonomy for this domain, best of our knowledge. Our results demonstrate that effective user clarification is essential for maximizing task success and conversational efficiency, minimizing unnecessary user interactions. The proposed taxonomy enhances accuracy in both single-agent and multi-agent settings, with the multi-agent approach yielding superior results due to effective task sharing with domain experts. Notably, MAC is model-agnostic and significantly boosts performance across both open-source and proprietary LLMs, with particularly pronounced gains for open-source models; helping close the performance gap through optimal design and supervision. Our ablation studies highlight that high-level ambiguity and vagueness handling by the supervisor is especially critical for robust performance in real-world scenarios. Overall, MAC establishes a foundation for future research and deployment of multi-agent, user-centric conversational systems, offering clear benefits for practical and industrial applications.

\paragraph{Limitations}
Although MAC is a new effective and promising framework, it has several limitations. First, due to the complexity of multi-turn conversations, successful goal-oriented dialogue tasks and planning require large-scale capable models. As a result, our experiments primarily rely on large-parameter, API-based models. Teaching these capabilities to smaller LLMs and deploying them efficiently remains an open challenge~\citep{belcak2025small}.
Second, we use LLM-as-a-Judge (LaaJ) based evaluations, which has known limitations in its core. In rare cases, the evaluator LLM may hallucinate and assign positive scores to incorrect dialogue trajectories; even though, we did not observe such failures, outlier cases may still exist. Nevertheless, LaaJ has been widely adopted in prior TOD research and has proven effective for evaluating complex, long-horizon conversational rollouts that are difficult to assess with deterministic metrics alone~\citep{xu2024autotod, acikgoz2025tdeval}. Combining LaaJ with complementary deterministic, rule-based metrics could improve robustness, but designing such hybrid evaluations remains an open challenge.
Finally, our error analysis reveals that in rare cases (approximately 1\%), the agent and the user simulator can become stuck in a loop~\citep{barres2025tau}, where the agent repeatedly asks clarification questions and the simulator returns the same response. Although infrequent, this behavior highlights the need for more realistic user simulators and points to human–agent co-evolution as an interesting direction.

\vspace{-0.75mm}

\paragraph{Future Work}
Handling user interactions in conversational agents is non-trivial, as it requires managing multiple tasks such as providing accurate responses or invoking the appropriate API from thousands of available tools~\citep{su2025toolorchestra}. In addition to these requirements, MAC further focus on challenge at scale through its carefully designed multi-agent setup, which is specifically tailored for user clarification. As a future direction, agents could learn optimal timing for seeking user clarification by monitoring environmental signals during interactions, leveraging recent reinforcement learning techniques~\citep{lambert2024tulu, guo2025deepseekr1} to continuously self-update and become increasingly successful over time as a promising path towards self-improving agents~\citep{zhang2025darwin, acikgoz2025ttsi}. Moreover, while our evaluation focuses on task success and the number of conversational turns as proxies for efficiency, quantifying overall user satisfaction remains an open question~\citep{terry2023interactive}, where factors such as dialogue naturalness and other user-centric elements may play a pivotal role.

\vspace{-2mm}

\bibliography{custom}

@inproceedings{aliannejadi2019asking,
  title={Asking clarifying questions in open-domain information-seeking conversations},
  author={Aliannejadi, Mohammad and Zamani, Hamed and Crestani, Fabio and Croft, W Bruce},
  booktitle={Proceedings of the 42nd international acm sigir conference on research and development in information retrieval},
  pages={475--484},
  year={2019}
}

@inproceedings{deng2023proactivedialoguesurvey,
  title={A survey on proactive dialogue systems: problems, methods, and prospects},
  author={Deng, Yang and Lei, Wenqiang and Lam, Wai and Chua, Tat-Seng},
  booktitle={Proceedings of the Thirty-Second International Joint Conference on Artificial Intelligence},
  pages={6583--6591},
  year={2023}
}

@article{sun2025multiagentappsurvey,
  title={Multi-agent coordination across diverse applications: A survey},
  author={Sun, Lijun and Yang, Yijun and Duan, Qiqi and Shi, Yuhui and Lyu, Chao and Chang, Yu-Cheng and Lin, Chin-Teng and Shen, Yang},
  journal={arXiv preprint arXiv:2502.14743},
  year={2025}
}

@inproceedings{
ong2025routellm,
title={Route{LLM}: Learning to Route {LLM}s from Preference Data},
author={Isaac Ong and Amjad Almahairi and Vincent Wu and Wei-Lin Chiang and Tianhao Wu and Joseph E. Gonzalez and M Waleed Kadous and Ion Stoica},
booktitle={The Thirteenth International Conference on Learning Representations},
year={2025},
url={https://openreview.net/forum?id=8sSqNntaMr}
}

@article{kuhn2022clam,
  title={Clam: Selective clarification for ambiguous questions with generative language models},
  author={Kuhn, Lorenz and Gal, Yarin and Farquhar, Sebastian},
  journal={arXiv preprint arXiv:2212.07769},
  year={2022}
}

@article{zhang2023clarify,
  title={Clarify when necessary: Resolving ambiguity through interaction with lms},
  author={Zhang, Michael JQ and Choi, Eunsol},
  journal={arXiv preprint arXiv:2311.09469},
  year={2023}
}

@article{andukuri2024stargate,
  title={Star-gate: Teaching language models to ask clarifying questions},
  author={Andukuri, Chinmaya and Fr{\"a}nken, Jan-Philipp and Gerstenberg, Tobias and Goodman, Noah D},
  journal={arXiv preprint arXiv:2403.19154},
  year={2024}
}

@article{zhang2024modeling,
  title={Modeling future conversation turns to teach llms to ask clarifying questions},
  author={Zhang, Michael JQ and Knox, W Bradley and Choi, Eunsol},
  journal={arXiv preprint arXiv:2410.13788},
  year={2024}
}

@article{dongre2024respact,
  title={ReSpAct: Harmonizing Reasoning, Speaking, and Acting Towards Building Large Language Model-Based Conversational AI Agents},
  author={Dongre, Vardhan and Yang, Xiaocheng and Acikgoz, Emre Can and Dey, Suvodip and Tur, Gokhan and Hakkani-T{\"u}r, Dilek},
  journal={arXiv preprint arXiv:2411.00927},
  year={2024}
}

@inproceedings{guo2024multiagentsurvey,
author = {Guo, Taicheng and Chen, Xiuying and Wang, Yaqi and Chang, Ruidi and Pei, Shichao and Chawla, Nitesh V. and Wiest, Olaf and Zhang, Xiangliang},
title = {Large language model based multi-agents: a survey of progress and challenges},
year = {2024},
isbn = {978-1-956792-04-1},
url = {https://doi.org/10.24963/ijcai.2024/890},
doi = {10.24963/ijcai.2024/890},
booktitle = {Proceedings of the Thirty-Third International Joint Conference on Artificial Intelligence},
articleno = {890},
numpages = {10},
location = {Jeju, Korea},
series = {IJCAI '24}
}

@inproceedings{ye2022multiwoz24,
    title = "{M}ulti{WOZ} 2.4: A Multi-Domain Task-Oriented Dialogue Dataset with Essential Annotation Corrections to Improve State Tracking Evaluation",
    author = "Ye, Fanghua  and
      Manotumruksa, Jarana  and
      Yilmaz, Emine",
    editor = "Lemon, Oliver  and
      Hakkani-Tur, Dilek  and
      Li, Junyi Jessy  and
      Ashrafzadeh, Arash  and
      Garcia, Daniel Hern{\'a}ndez  and
      Alikhani, Malihe  and
      Vandyke, David  and
      Du{\v{s}}ek, Ond{\v{r}}ej",
    booktitle = "Proceedings of the 23rd Annual Meeting of the Special Interest Group on Discourse and Dialogue",
    month = sep,
    year = "2022",
    address = "Edinburgh, UK",
    publisher = "Association for Computational Linguistics",
    url = "https://aclanthology.org/2022.sigdial-1.34/",
    doi = "10.18653/v1/2022.sigdial-1.34",
    pages = "351--360"
}

@article{li2023asktoclarify,
  title={Eliciting human preferences with language models},
  author={Li, Belinda Z and Tamkin, Alex and Goodman, Noah and Andreas, Jacob},
  journal={arXiv preprint arXiv:2310.11589},
  year={2023}
}

@inproceedings{zhang2025clarifywhennecessary,
    title = "Clarify When Necessary: Resolving Ambiguity Through Interaction with {LM}s",
    author = "Zhang, Michael JQ  and
      Choi, Eunsol",
    editor = "Chiruzzo, Luis  and
      Ritter, Alan  and
      Wang, Lu",
    booktitle = "Findings of the Association for Computational Linguistics: NAACL 2025",
    month = apr,
    year = "2025",
    address = "Albuquerque, New Mexico",
    publisher = "Association for Computational Linguistics",
    url = "https://aclanthology.org/2025.findings-naacl.306/",
    doi = "10.18653/v1/2025.findings-naacl.306",
    pages = "5526--5543",
    ISBN = "979-8-89176-195-7"
}

@article{tran2025multiagentmechnaism,
  title={Multi-Agent Collaboration Mechanisms: A Survey of LLMs},
  author={Tran, Khanh-Tung and Dao, Dung and Nguyen, Minh-Duong and Pham, Quoc-Viet and O'Sullivan, Barry and Nguyen, Hoang D},
  journal={arXiv preprint arXiv:2501.06322},
  year={2025}
}

@article{hou2025mcp,
  title={Model context protocol (mcp): Landscape, security threats, and future research directions},
  author={Hou, Xinyi and Zhao, Yanjie and Wang, Shenao and Wang, Haoyu},
  journal={arXiv preprint arXiv:2503.23278},
  year={2025}
}

@inproceedings{ehsan2020simpletod,
 author = {Hosseini-Asl, Ehsan and McCann, Bryan and Wu, Chien-Sheng and Yavuz, Semih and Socher, Richard},
 booktitle = {Advances in Neural Information Processing Systems},
 editor = {H. Larochelle and M. Ranzato and R. Hadsell and M.F. Balcan and H. Lin},
 pages = {20179--20191},
 publisher = {Curran Associates, Inc.},
 title = {A Simple Language Model for Task-Oriented Dialogue},
 url = {https://proceedings.neurips.cc/paper_files/paper/2020/file/e946209592563be0f01c844ab2170f0c-Paper.pdf},
 volume = {33},
 year = {2020}
}

@inproceedings{yang2021ubar,
  title={Ubar: Towards fully end-to-end task-oriented dialog system with gpt-2},
  author={Yang, Yunyi and Li, Yunhao and Quan, Xiaojun},
  booktitle={Proceedings of the AAAI conference on artificial intelligence},
  volume={35},
  number={16},
  pages={14230--14238},
  year={2021}
}

@inproceedings{zhong2023galaxy,
  title={A Task-oriented Dialog Model with Task-progressive and Policy-aware Pre-training},
  author={Zhong, Lucen and Lu, Hengtong and Yuan, Caixia and Wang, Xiaojie and Sun, Jiashen and Zeng, Ke and Wan, Guanglu},
  booktitle={CCF International Conference on Natural Language Processing and Chinese Computing},
  pages={3--15},
  year={2023},
  organization={Springer}
}

@inproceedings{sun2023mars,
    title = "{M}ars: Modeling Context {\&} State Representations with Contrastive Learning for End-to-End Task-Oriented Dialog",
    author = "Sun, Haipeng  and
      Bao, Junwei  and
      Wu, Youzheng  and
      He, Xiaodong",
    editor = "Rogers, Anna  and
      Boyd-Graber, Jordan  and
      Okazaki, Naoaki",
    booktitle = "Findings of the Association for Computational Linguistics: ACL 2023",
    month = jul,
    year = "2023",
    address = "Toronto, Canada",
    publisher = "Association for Computational Linguistics",
    url = "https://aclanthology.org/2023.findings-acl.708/",
    doi = "10.18653/v1/2023.findings-acl.708",
    pages = "11139--11160"
}

@inproceedings{bang2023toatod,
    title = "Task-Optimized Adapters for an End-to-End Task-Oriented Dialogue System",
    author = "Bang, Namo  and
      Lee, Jeehyun  and
      Koo, Myoung-Wan",
    editor = "Rogers, Anna  and
      Boyd-Graber, Jordan  and
      Okazaki, Naoaki",
    booktitle = "Findings of the Association for Computational Linguistics: ACL 2023",
    month = jul,
    year = "2023",
    address = "Toronto, Canada",
    publisher = "Association for Computational Linguistics",
    url = "https://aclanthology.org/2023.findings-acl.464/",
    doi = "10.18653/v1/2023.findings-acl.464",
    pages = "7355--7369"
}

@inproceedings{li2024fnctod,
    title = "Large Language Models as Zero-shot Dialogue State Tracker through Function Calling",
    author = "Li, Zekun  and
      Chen, Zhiyu  and
      Ross, Mike  and
      Huber, Patrick  and
      Moon, Seungwhan  and
      Lin, Zhaojiang  and
      Dong, Xin  and
      Sagar, Adithya  and
      Yan, Xifeng  and
      Crook, Paul",
    editor = "Ku, Lun-Wei  and
      Martins, Andre  and
      Srikumar, Vivek",
    booktitle = "Proceedings of the 62nd Annual Meeting of the Association for Computational Linguistics (Volume 1: Long Papers)",
    month = aug,
    year = "2024",
    address = "Bangkok, Thailand",
    publisher = "Association for Computational Linguistics",
    url = "https://aclanthology.org/2024.acl-long.471/",
    doi = "10.18653/v1/2024.acl-long.471",
    pages = "8688--8704"
}

@inproceedings{xu2024autotod,
    title = "Rethinking Task-Oriented Dialogue Systems: From Complex Modularity to Zero-Shot Autonomous Agent",
    author = "Xu, Heng-Da  and
      Mao, Xian-Ling  and
      Yang, Puhai  and
      Sun, Fanshu  and
      Huang, Heyan",
    editor = "Ku, Lun-Wei  and
      Martins, Andre  and
      Srikumar, Vivek",
    booktitle = "Proceedings of the 62nd Annual Meeting of the Association for Computational Linguistics (Volume 1: Long Papers)",
    month = aug,
    year = "2024",
    address = "Bangkok, Thailand",
    publisher = "Association for Computational Linguistics",
    url = "https://aclanthology.org/2024.acl-long.152/",
    doi = "10.18653/v1/2024.acl-long.152",
    pages = "2748--2763"
}

@inproceedings{hudecek2023arellms,
    title = "Are Large Language Models All You Need for Task-Oriented Dialogue?",
    author = "Hude{\v{c}}ek, Vojt{\v{e}}ch  and
      Dusek, Ondrej",
    editor = "Stoyanchev, Svetlana  and
      Joty, Shafiq  and
      Schlangen, David  and
      Dusek, Ondrej  and
      Kennington, Casey  and
      Alikhani, Malihe",
    booktitle = "Proceedings of the 24th Annual Meeting of the Special Interest Group on Discourse and Dialogue",
    month = sep,
    year = "2023",
    address = "Prague, Czechia",
    publisher = "Association for Computational Linguistics",
    url = "https://aclanthology.org/2023.sigdial-1.21/",
    doi = "10.18653/v1/2023.sigdial-1.21",
    pages = "216--228"
}

@inproceedings{hu2022icl,
    title = "In-Context Learning for Few-Shot Dialogue State Tracking",
    author = "Hu, Yushi  and
      Lee, Chia-Hsuan  and
      Xie, Tianbao  and
      Yu, Tao  and
      Smith, Noah A.  and
      Ostendorf, Mari",
    editor = "Goldberg, Yoav  and
      Kozareva, Zornitsa  and
      Zhang, Yue",
    booktitle = "Findings of the Association for Computational Linguistics: EMNLP 2022",
    month = dec,
    year = "2022",
    address = "Abu Dhabi, United Arab Emirates",
    publisher = "Association for Computational Linguistics",
    url = "https://aclanthology.org/2022.findings-emnlp.193/",
    doi = "10.18653/v1/2022.findings-emnlp.193",
    pages = "2627--2643"
}

@inproceedings{chung2023instructtod,
    title = "{I}nstruct{TODS}: Large Language Models for End-to-End Task-Oriented Dialogue Systems",
    author = "Chung, Willy  and
      Cahyawijaya, Samuel  and
      Wilie, Bryan  and
      Lovenia, Holy  and
      Fung, Pascale",
    editor = "Chen, Kehai  and
      Ku, Lun-Wei",
    booktitle = "Proceedings of the Second Workshop on Natural Language Interfaces",
    month = nov,
    year = "2023",
    address = "Bali, Indonesia",
    publisher = "Association for Computational Linguistics",
    url = "https://aclanthology.org/2023.nlint-1.1/",
    doi = "10.18653/v1/2023.nlint-1.1",
    pages = "1--21"
}

@inproceedings{
gupta2024dard,
title={{DARD}: A Multi-Agent Approach for Task-Oriented Dialog Systems},
author={Aman Gupta and Anirudh Ravichandran and Narayanan Sadagopan and Anurag Beniwal},
booktitle={NeurIPS 2024 Workshop on Open-World Agents},
year={2024},
url={https://openreview.net/forum?id=RbkX9e4qqP}
}

@inproceedings{acikgoz2025coalm,
    title = "Can a Single Model Master Both Multi-turn Conversations and Tool Use? {C}o{ALM}: A Unified Conversational Agentic Language Model",
    author = {Acikgoz, Emre Can  and
      Greer, Jeremiah  and
      Datta, Akul  and
      Yang, Ze  and
      Zeng, William  and
      Elachqar, Oussama  and
      Koukoumidis, Emmanouil  and
      Hakkani-T{\"u}r, Dilek  and
      Tur, Gokhan},
    editor = "Che, Wanxiang  and
      Nabende, Joyce  and
      Shutova, Ekaterina  and
      Pilehvar, Mohammad Taher",
    booktitle = "Proceedings of the 63rd Annual Meeting of the Association for Computational Linguistics (Volume 1: Long Papers)",
    month = jul,
    year = "2025",
    address = "Vienna, Austria",
    publisher = "Association for Computational Linguistics",
    url = "https://aclanthology.org/2025.acl-long.605/",
    doi = "10.18653/v1/2025.acl-long.605",
    pages = "12370--12390",
    ISBN = "979-8-89176-251-0"
}

@article{hurst2024gpt4o,
  title={Gpt-4o system card},
  author={Hurst, Aaron and Lerer, Adam and Goucher, Adam P and Perelman, Adam and Ramesh, Aditya and Clark, Aidan and Ostrow, AJ and Welihinda, Akila and Hayes, Alan and Radford, Alec and others},
  journal={arXiv preprint arXiv:2410.21276},
  year={2024}
}

@article{yang2025qwen3,
  title={Qwen3 technical report},
  author={Yang, An and Li, Anfeng and Yang, Baosong and Zhang, Beichen and Hui, Binyuan and Zheng, Bo and Yu, Bowen and Gao, Chang and Huang, Chengen and Lv, Chenxu and others},
  journal={arXiv preprint arXiv:2505.09388},
  year={2025}
}

@article{guo2025deepseekr1,
  title={Deepseek-r1: Incentivizing reasoning capability in llms via reinforcement learning},
  author={Guo, Daya and Yang, Dejian and Zhang, Haowei and Song, Junxiao and Zhang, Ruoyu and Xu, Runxin and Zhu, Qihao and Ma, Shirong and Wang, Peiyi and Bi, Xiao and others},
  journal={arXiv preprint arXiv:2501.12948},
  year={2025}
}

@article{acikgoz2025desideratum,
  title={A desideratum for conversational agents: Capabilities, challenges, and future directions},
  author={Acikgoz, Emre Can and Qian, Cheng and Wang, Hongru and Dongre, Vardhan and Chen, Xiusi and Ji, Heng and Hakkani-T{\"u}r, Dilek and Tur, Gokhan},
  journal={arXiv preprint arXiv:2504.16939},
  year={2025}
}

@inproceedings{acikgoz2025tdeval,
    title = "{TD}-{EVAL}: Revisiting Task-Oriented Dialogue Evaluation by Combining Turn-Level Precision with Dialogue-Level Comparisons",
    author = "Acikgoz, Emre Can  and
      Guo, Carl  and
      Dey, Suvodip  and
      Datta, Akul  and
      Kim, Takyoung  and
      Tur, Gokhan  and
      Hakkani-Tur, Dilek",
    editor = "B{\'e}chet, Fr{\'e}d{\'e}ric  and
      Lef{\`e}vre, Fabrice  and
      Asher, Nicholas  and
      Kim, Seokhwan  and
      Merlin, Teva",
    booktitle = "Proceedings of the 26th Annual Meeting of the Special Interest Group on Discourse and Dialogue",
    month = aug,
    year = "2025",
    address = "Avignon, France",
    publisher = "Association for Computational Linguistics",
    url = "https://aclanthology.org/2025.sigdial-1.7/",
    pages = "113--132"
}

@ARTICLE{acikgoz2025todllms,
  author={Acikgoz, Emre Can and Hakkani-Tür, Dilek and Tur, Gokhan},
  journal={IEEE Signal Processing Magazine}, 
  title={Conversational Agents in the Era of Large Language Models [Perspectives]}, 
  year={2025},
  volume={42},
  number={3},
  pages={35-39},
  doi={10.1109/MSP.2025.3565479}}

@article{belcak2025small,
  title={Small Language Models are the Future of Agentic AI},
  author={Belcak, Peter and Heinrich, Greg and Diao, Shizhe and Fu, Yonggan and Dong, Xin and Muralidharan, Saurav and Lin, Yingyan Celine and Molchanov, Pavlo},
  journal={arXiv preprint arXiv:2506.02153},
  year={2025}
}

@article{zhang2025darwin,
  title={Darwin Godel Machine: Open-Ended Evolution of Self-Improving Agents},
  author={Zhang, Jenny and Hu, Shengran and Lu, Cong and Lange, Robert and Clune, Jeff},
  journal={arXiv preprint arXiv:2505.22954},
  year={2025}
}

@article{acikgoz2025ttsi,
  title={Self-improving llm agents at test-time},
  author={Acikgoz, Emre Can and Qian, Cheng and Ji, Heng and Hakkani-T{\"u}r, Dilek and Tur, Gokhan},
  journal={arXiv preprint arXiv:2510.07841},
  year={2025}
}

@article{su2025toolorchestra,
  title={ToolOrchestra: Elevating Intelligence via Efficient Model and Tool Orchestration},
  author={Su, Hongjin and Diao, Shizhe and Lu, Ximing and Liu, Mingjie and Xu, Jiacheng and Dong, Xin and Fu, Yonggan and Belcak, Peter and Ye, Hanrong and Yin, Hongxu and others},
  journal={arXiv preprint arXiv:2511.21689},
  year={2025}
}

@article{lambert2024tulu,
  title={Tulu 3: Pushing frontiers in open language model post-training},
  author={Lambert, Nathan and Morrison, Jacob and Pyatkin, Valentina and Huang, Shengyi and Ivison, Hamish and Brahman, Faeze and Miranda, Lester James V and Liu, Alisa and Dziri, Nouha and Lyu, Shane and others},
  journal={arXiv preprint arXiv:2411.15124},
  year={2024}
}

@misc{barres2025tau,
      title={$\tau^2$-Bench: Evaluating Conversational Agents in a Dual-Control Environment}, 
      author={Victor Barres and Honghua Dong and Soham Ray and Xujie Si and Karthik Narasimhan},
      year={2025},
      eprint={2506.07982},
      archivePrefix={arXiv},
      primaryClass={cs.AI},
      url={https://arxiv.org/abs/2506.07982}, 
}

@article{terry2023interactive,
  title={Interactive AI alignment: Specification, process, and evaluation alignment},
  author={Terry, Michael and Kulkarni, Chinmay and Wattenberg, Martin and Dixon, Lucas and Morris, Meredith Ringel},
  journal={arXiv preprint arXiv:2311.00710},
  year={2023}
}

\clearpage
\appendix
\section{Further Details on Evaluation and MultiWOZ 2.4 } 
\label{app:evaluation}

We evaluate the performance of our MAC framework using dialogue-level metrics that capture both the effectiveness and efficiency of task completion. Our primary metric is Success Rate, which measures whether the agent fully satisfies all user-specified constraints and successfully completes the task. For each dialogue, we use an LLM-based judge using \texttt{GPT-4o-2024-11-20}~\citep{hurst2024gpt4o} to assess if the agent’s final response fulfills every requirement defined by the user’s goal, including both requested attributes (such as hotel name or train arrival time) and booking constraints (such as the number of people or destination) following \citet{xu2024autotod}. Formally, a dialogue is considered successful if all constraints in the user’s goal $G$ are met by the end of the interaction:
\begin{equation}
\text{Success} = \mathbb{I}(\text{all constraints in } G \text{ are satisfied}),
\end{equation}
where $\mathbb{I}(\cdot)$ denotes the indicator function. This score is computed for every dialogue and averaged across the evaluation set. To account for the stochastic nature of both model inference and LLM-based judging, we conduct five independent runs for each experimental configuration. 

We report two aggregate Success Rate metrics: \textbf{Success Max@5}, the highest single-run success rate out of five runs, reflecting the best-case performance; and \textbf{Success Avg@5}, the mean and standard deviation of success rates over the five runs, providing a robust measure of typical performance and variance. In addition, we report the \textbf{Average Number of Turns} per conversation as an efficiency metric. This measures the average length of the dialogue required to complete the task, with lower values indicating more concise and effective interactions. This metric is particularly important for assessing the practical impact of clarification strategies on user burden and overall system efficiency.

\begin{table*}[h!]
\centering
\resizebox{\linewidth}{!}{%
\begin{tabular}{lrcc}
\toprule
\textbf{Domain} & \textbf{API Name}   & \textbf{API Arguments}                                          & \textbf{Test Samples per Domain}   \\ \midrule
Restaurant      & query\_restaurant   & area, pricerange, food, name                                    & 437   \\               
                & book\_restaurant    & name, people, day, time, pricerange, stars, type                &    \\ \midrule
Hotel           & query\_hotel        & area, internet, name, parking                                   & 394        \\
                & book\_hotel         & name, people, day, stay                                         &          \\ \midrule
Attraction      & query\_attraction   & area, name, type                                                & 395   \\ \midrule
Train           & query\_train        & arriveBy, day, departure, destination, leaveAt, trainID         & 494       \\
                & buy\_train\_ticket  & arriveBy, day, departure, destination, leaveAt, trainID, people &   \\ \midrule
Taxi            & book\_taxi          & arriveBy, departure, destination, leaveAt                       & 195          \\ \bottomrule
\end{tabular}%
}
\caption{Available actions per domain in MultiWOZ 2.4 dataset.}
\label{tab:dataset}
\end{table*}
\begin{figure*}[t]
\centering
\begin{tcolorbox}[title=ClarifyRLData Generation Prompt]
\small
Given the following User Goal and Ground Truth Conversation, update the conversation to introduce ambiguity or underspecification in the user’s turns, such that the agent must ask for clarification \textbf{at least once and at most three times}. For every agent clarification, enclose the clarification in `\texttt{<clarify>}` and `\texttt{</clarify>}` tokens. After each agent clarification, update the following user turn(s) to resolve the ambiguity.\\
Do not change the overall goal or successful task completion. Only modify the conversation for clarification needs. \\
\\
\textbf{Task:}\\
- Carefully read the User Goal and the Ground Truth Conversation.\\
- Rewrite the conversation so that some user turns are ambiguous or missing key information, requiring the agent to clarify at least once and at most three times (with the `\texttt{<clarify>}` tokens).\\
- Keep the rest of the conversation as natural as possible and ensure the final output still accomplishes the user goal.\\
\\
\textbf{Output Format:}\\
\\
- Output the \textbf{updated conversation only} as a \textbf{valid JSON array} in the format:\\
\\
```json\\
\{\\
\hspace*{2em}{"from": "user", "value": "..."},\\
\hspace*{2em}{"from": "agent", "value": "\texttt{<clarify>}...\texttt{</clarify>}"},\\
\hspace*{2em}{"from": "user", "value": "..."},\\
\hspace*{2em}{"from": "agent", "value": "..."},\\
\hspace*{1em}...\\
\}\\
```\\
\\
- \textbf{No extra text, no comments, no explanations, no markdown—just the JSON array.} \\
- The output \textbf{must} be valid JSON. \\
- If the format is not exactly correct, data loading will fail. \\
\\
\textbf{User Goal}\\
\texttt{<user\_goal>}\\
\\
\textbf{Ground Truth Conversation}\\
\texttt{<conversation>}\\
\\
\textbf{New Conversation with User Clarification} \\
\{Your JSON Data Here\}\\

\end{tcolorbox}
\caption{\textbf{ClarifyRL Data Generation Prompt.} Prompt used to synthesize ClarifyRL data with an LLM, conditioned on a given user goal and dialogue from the MultiWOZ 2.4 training split.}
\label{fig:user_satisfaction_prompt}
\end{figure*}

\section{Experimental Details}
During our experiments, we use the OpenAI API\footnote{\url{https://openai.com/api/}} to evaluate the GPT models \texttt{GPT-4o-2024-11-20}~\citep{hurst2024gpt4o} and \texttt{gpt-4o-mini}. For open-source model evaluations, we use the TogetherAI API\footnote{\url{https://api.together.xyz/}} with \texttt{Qwen3-235B-A22B}~\citep{yang2025qwen3}. To ensure reproducibility, we use default generation settings for all models without tuning any inference hyperparameters. In terms of runtime, a single model inference takes approximately 15 minutes, while evaluation requires around 4–5 minutes. Since LLM-based judge metrics are non-deterministic, the results may vary slightly across runs. To account for this variability, we perform all evaluations five times and report scores with standard deviations.



\begin{figure*}[t]
\centering

\begin{tcolorbox}[title=Supervisor Agent Prompt]
\small

\textbf{\#\#\# Task} \\
You are an expert routing agent in a multi-domain conversational AI system for the MultiWOZ dataset. 
Your specific task is to analyze a user's query and determine which of the following five domain experts is best suited to handle it:\\
\hspace*{2em} 1. restaurant (for queries related to finding, booking, or getting information about restaurants) \\
\hspace*{2em} 2. hotel (for queries related to finding, booking, or getting information about hotels or other accommodations)\\
\hspace*{2em} 3. attraction (for queries related to finding or getting information about tourist attractions, landmarks, or points of interest)\\
\hspace*{2em} 4. train (for queries related to finding, booking, or getting information about train travel)\\
\hspace*{2em} 5. taxi (for queries related to booking or getting information about taxi services)\\
\\
Read the user's query provided below (\#\#\# User Query). Your goal is to identify the single, most dominant domain relevant to the query.\\
\\
\textbf{\#\#\# Output Instructions} \\
You MUST output ONLY the exact lowercase label corresponding to the selected domain, enclosed in \texttt{<domain>} and \texttt{</domain>}. \\
For example, if the query is about a hotel, your output must be \texttt{<domain>}hotel\texttt{</domain>}.
Do NOT include any other words, phrases, explanations, or punctuation outside of these tags. Your entire response should be just one of these five labels, wrapped in the domain tags as shown below:\\
\\
\texttt{<domain>}restaurant\texttt{</domain>}\\
\texttt{<domain>}hotel\texttt{</domain>}\\
\texttt{<domain>}attraction\texttt{</domain>}\\
\texttt{<domain>}train\texttt{</domain>}\\
\texttt{<domain>}taxi\texttt{</domain>}\\
\\
If the query seems to touch on multiple domains, select the one that appears to be the primary focus or the one that needs to be addressed first. \\
If no domain is clearly identifiable from the list, you must still choose the closest possible one or a default agreed upon (though for this specific instruction, you must pick one of the five and wrap it in the tags).\\
\\
\textbf{\#\#\# User Query} \\
\{\{user\_query\}\}\\
\\
Selected Domain Label:\\

\end{tcolorbox}
\caption{\textbf{LLM Prompt} used for the Supervisor agent to decide domain routing.}
\label{fig:supervisor-prompt}
\end{figure*}
\begin{figure*}[t]
\centering

\begin{tcolorbox}[title=Hotel Domain Expert Prompt]
\small

\textbf{\#\#\# Role Description} \\
You are an advanced AI assistant specializing in conversational dialogues focused on the hotel domain.  \\
You can act both as a system (providing hotel information and booking services) and a user (interacting with the hotel database) to assist users in completing hotel-related tasks. \\
 \\
\textbf{\#\#\# Task Information} \\
- Each time, you must determine whether to call an API by reasoning through "Thought:". \\
- If you decide that an API call is necessary, include "Thought:" for reasoning, followed by "API Name:", "API Input:", "API Result:". \\
- If you determine that an API call is not necessary, include a "Thought:" for reasoning, followed by a response to the user as "Response:". \\
- If the user asks for some attributes of a hotel (e.g., address, phone number, price range, parking, internet), then an API call is necessary. \\
- You are not allowed to use APIs not mentioned below. If you decide that the mentioned APIs are not sufficient for the user's request, you should inform the user that you can only assist with hotel queries and bookings. \\
- If you decide that more than one API call is needed (e.g., query first, then book), you should call one API first and wait for the API result. After obtaining that result, you may think and call the next API or think and make a response. \\
- The user can sometimes not care about the value of an API input slot and may mention it explicitly in the conversation (e.g., "I don't care about the price range"). In such cases, predict "dontcare" as a slot value for that particular slot. \\
- If you decide that there is an API input slot that the user has never mentioned and is required for the API, please put "any" as the slot value as a placeholder. \\
- You can put only one value in each API input slot per query. \\
 \\
ATTENTION: \\
- Predict "dontcare" as a slot value ONLY if the user has explicitly mentioned it in the conversation. \\
 \\
\textbf{\#\#\# Output Format} \\
- If an API Call is Needed: \\
\hspace*{2em}     Thought: [Your reasoning for why an API call is needed] \\
\hspace*{2em}     API Name: [Available APIs: query\_hotels, book\_hotel] \\
\hspace*{2em}     API Input: [The input parameters for the API as a JSON] \\
\hspace*{2em}     API Result: \\
 \\
- If an API Call is Not Needed: \\
\hspace*{2em}     Thought: [Your reasoning for why an API call is not needed and you are responding directly] \\
\hspace*{2em}     Response: [Your response to the user] \\
 \\
\textbf{\#\#\# API Details:} \\
\{\{api\_descriptions\}\}
 \\
\textbf{\#\#\# Example with explanation} \\
\{\{example\_conversation\}\}
 
\end{tcolorbox}
\caption{\textbf{LLM Prompt} used for the Domain Expert agent to decide response.}
\label{fig:expert-prompt}
\end{figure*}

\begin{figure*}[t]
\centering
\begin{tcolorbox}[title=Supervisor Agent Prompt with User Clarification]
\small
\textbf{\#\#\# Task}\\
You are a high-level supervisor and routing agent in a multi-domain conversational AI system. Your primary goal is to analyze a user's query and take one of two actions:\\
1.  \textbf{Clarify:} If the user's intent or desired domain is ambiguous, ask a single, precise clarification question.\\
2.  \textbf{Route:} If the user's intent and domain are clear, route the query to the appropriate expert agent.\\
\\
\textbf{\#\#\# Core Directive}\\
Your responsibility is to handle ONLY high-level, common-sense, and domain-agnostic ambiguities. \textbf{DO NOT} ask for domain-specific details (e.g., cuisine, price range, number of people, time of booking). Your task is to figure out \textbf{WHAT} the user wants in general, not the specifics of how to do it.\\
\\
\textbf{\#\#\# Clarification Taxonomy (When to Ask)}\\
Before routing, you must determine if clarification is needed. Use this taxonomy to guide your decision:\\
-   \textbf{Domain Ambiguity:} The query could fit multiple domains (e.g., "Find me a good place." - good place to eat or stay?).\\
-   \textbf{Intent Ambiguity:} The domain is clear, but the user's goal is not (e.g., "Tell me about trains" - find a schedule or book a ticket?).\\
-   \textbf{Vague Goal Specification:} The query is too broad to be actionable (e.g., "Help me with my trip" - what kind of help; booking or search?).\\
-   \textbf{Contextual Disambiguation:} The query uses vague references like "it" or "that place" which are unclear from the context.\\
-   \textbf{General Conflict:} Broad contradictions in user's input not domain-specific (e.g., "I changed my mind about the date").\\
-   \textbf{General Noise/Correction:} Common errors or typos needing clarification (e.g., "I meant tomorrow not today").\\
-   \textbf{Unfamiliar Domain Request:} Request does not match known domains clearly (e.g., "Can you help me fix my phone?" - no such a domain or expert).\\
- If a clarification is needed, \textbf{always output a clarifying question in the format:}\\
    \hspace*{2em}     - Thought: The user request is unclear due to [reason].\\
    \hspace*{2em}     - Response: \texttt{<clarify>}[Your response question to the user for clarification]\texttt{</clarify>}\\
\\
\textbf{\#\#\# Output Instructions}\\
1. For user clarification, you should provide your reasoning as "Thought: [your reasoning]" and your user clarification question response as "Response: \texttt{<clarify>}[your high-level clarification question]\texttt{</clarify>}".\\
2. For domain selection, you should provide your response between \texttt{<domain>} and \texttt{</domain>} tags.\\
\\
Example: \\
\\
\textbf{\#\#\# Output Format}\\
You MUST output either a `\texttt{<clarify>}` tag OR a `\texttt{<domain>}` tag in a single turn.\\
\\
\textbf{1. If User Clarification is Needed:}\\
    \hspace*{2em}     Thought: [Your reasoning for why user clarification is needed, you are responding directly]\\
    \hspace*{2em}     Response: \texttt{<clarify>}[Your response question to the user]\texttt{</clarify>}\\
\\
\textbf{2.  If The Query is Clear and Routable:}\\
    \hspace*{2em}     Your entire output must be the single, lowercase domain label wrapped in the `\texttt{<domain>}` tag.\\
   \hspace*{2em}      You MUST output ONLY the exact lowercase label corresponding to the selected domain, enclosed in \texttt{<domain>} and \texttt{</domain>}. \\
    \hspace*{2em}     For example, if the query is about a hotel, your output must be \texttt{<domain>}hotel\texttt{</domain>}.Your entire response should be just one of these five labels, wrapped in the domain tags as shown below:\\
    \hspace*{2em}     -   \texttt{<domain>}restaurant\texttt{</domain>}\\
    \hspace*{2em}     -   \texttt{<domain>}hotel\texttt{</domain>}\\
    \hspace*{2em}     -   \texttt{<domain>}attraction\texttt{</domain>}\\
    \hspace*{2em}     -   \texttt{<domain>}train\texttt{</domain>}\\
    \hspace*{2em}     -   \texttt{<domain>}taxi\texttt{</domain>}\\
\\
If the query seems to touch on multiple domains, select the one that appears to be the primary focus or the one that needs to be addressed first.
If no domain is clearly identifiable from the list, you must still choose the closest possible one or a default agreed upon (though for this specific instruction, you must pick one of the five and wrap it in the tags).\\
\\
\textbf{\#\#\# Conversation History}\\
{{conversation\_history}}\\
\\
\textbf{\#\#\# User Query}\\
{{user\_query}}\\
\\
\textbf{\#\#\# Output:}\\
\end{tcolorbox}
\caption{\textbf{LLM Prompt} used for the Supervisor agent to decide domain routing and user clarification.}
\label{fig:mac-advisor}
\end{figure*}

\begin{figure*}[t]
\centering
\begin{tcolorbox}[title=Hotel Expert Prompt with User Clarification]
\small
\textbf{\#\#\# Role Description} \\
You are an advanced AI assistant specializing in conversational dialogues focused on the hotel domain.  \\
You can act both as a system (providing hotel information and booking services) and a user (interacting with the hotel database) to assist users in completing hotel-related tasks. \\
 \\
\textbf{\#\#\# Task Information} \\
- Each time, you must determine whether to call an API by reasoning through "Thought:". \\
- If you decide that an API call is necessary, include "Thought:" for reasoning, followed by "API Name:", "API Input:", "API Result:". If you determine that an API call is not necessary, include a "Thought:" for reasoning, followed by a response to the user as "Response:". \\
- If the user asks for some attributes of a hotel (e.g., address, phone number, price range, parking, internet), then an API call is necessary. \\
- You are not allowed to use APIs not mentioned below. If you decide that the mentioned APIs are not sufficient for the user's request, you should inform the user that you can only assist with hotel queries and bookings. \\
- If you decide that more than one API call is needed (e.g., query first, then book), you should call one API first and wait for the API result. After obtaining that result, you may think and call the next API or think and make a response. The user can sometimes not care about the value of an API input slot and may mention it explicitly in the conversation (e.g., "I don't care about the price range"). In such cases, predict "dontcare" as a slot value for that particular slot. \\
- If you decide that there is an API input slot that the user has never mentioned and is required for the API, please put "any" as the slot value as a placeholder. You can put only one value in each API input slot per query. \\
 \\
\textbf{\#\#\# Clarification Taxonomy (When to Ask)}\\
Before calling an API, determine if you have all the necessary information. If not, ask a clarifying question using this taxonomy:\\
-   \textbf{Parameter Underspecification:} Key details for a search or booking are missing (e.g., location, cuisine, number of people, time).\\
-   \textbf{Value Ambiguity/Vagueness:} A user's term is subjective and needs clarification (e.g., "a nice place," "somewhere soon").\\
-   \textbf{Constraint Conflict: }The user provides contradictory information (e.g., "a cheap but expensive restaurant").\\
-   \textbf{Entity Disambiguation/Not Found:} A specific restaurant name is ambiguous or cannot be found.\\
-   \textbf{Confirmation of Inferred Information:} You have inferred a detail from context and need to confirm it before proceeding.\\
- It is also important not to burden the user with repetitive or similar clarification questions in your overall conversation; please be mindful of this during your conversation.\\
- If a clarification is needed, \textbf{always output a clarifying question in the format:}\\
\hspace*{2em}    - Thought: The user request is unclear due to [reason].\\
\hspace*{2em}    - Response: \texttt{<clarify>}[Your response question to the user for clarification]\texttt{</clarify>}\\
- If you decided to ask the user for further clarification about the user query, you should output your user clarification question as: \texttt{<clarify>}...\texttt{</clarify>}. Your output should be like this:\\
\\
Thought: [Your reasoning for asking user clarification questions]\\
Response: \texttt{<clarify>}YOUR RESPONSE FOR USER CLARIFICATION HERE\texttt{</clarify>}\\
 \\
\textbf{\#\#\# Output Format} \\
- If an API Call is Needed: \\
\hspace*{2em}     Thought: [Your reasoning for why an API call is needed] \\
\hspace*{2em}     API Name: [Available APIs: query\_hotels, book\_hotel] \\
\hspace*{2em}     API Input: [The input parameters for the API as a JSON] \\
\hspace*{2em}     API Result: \\
 \\
- If an API Call is Not Needed: \\
\hspace*{2em}     Thought: [Your reasoning for why an API call is not needed and you are responding directly] \\
\hspace*{2em}     Response: [Your response to the user] \\
 \\
\textbf{\#\#\# API Details:} \\
\{\{api\_descriptions\}\}
\\
\textbf{\#\#\# Example with explanation} \\
\{\{example\_conversation\}\}
 
\end{tcolorbox}
\caption{\textbf{LLM Prompt} used for the Supervisor agent to decide domain routing and user clarification.}
\label{fig:mac-expert}
\end{figure*}

\end{document}